\documentclass[10pt,twocolumn,letterpaper]{article}

\usepackage[pagenumbers]{cvpr} 

\usepackage{float}
\definecolor{cvprblue}{rgb}{0.21,0.49,0.74}
\usepackage[pagebackref,breaklinks,colorlinks,allcolors=cvprblue]{hyperref}

\def\paperID{*****} 
\def\confName{CVPR}
\def\confYear{2025}

\title{Light Future: Multimodal Action Frame Prediction via InstructPix2Pix}

\author{
Zesen Zhong \quad Duomin Zhang \quad Yijia Li\\
School of Data Science,\\
The Chinese University of Hong Kong, Shenzhen\\
\texttt{\{zesenzhong,duominzhang,yijiali\}@link.cuhk.edu.cn}
}

\begin{document}

\maketitle
\begin{abstract}

Predicting future motion trajectories is a critical capability across domains such as robotics, autonomous systems, and human activity forecasting, enabling safer and more intelligent decision-making. This paper proposes a novel, efficient, and lightweight approach for robot action prediction, offering significantly reduced computational cost and inference latency compared to conventional video prediction models. Importantly, it pioneers the adaptation of the InstructPix2Pix model for forecasting future visual frames in robotic tasks, extending its utility beyond static image editing.

We implement a deep learning-based visual prediction framework that forecasts what a robot will observe 100 frames (10 seconds) into the future, given a current image and a textual instruction.
We innovatively repurpose and fine-tune the InstructPix2Pix model to accept both visual and textual inputs, enabling multimodal future frame prediction. Experiments on the RoboTWin dataset (generated based on real-world scenarios) demonstrate that our method achieves superior SSIM and PSNR compared to state-of-the-art baselines in robot action prediction tasks.

Unlike conventional video prediction models that require multiple input frames, heavy computation, and slow inference latency, our approach only needs a single image and a text prompt as input. This lightweight design enables faster inference, reduced GPU demands, and flexible multimodal control—particularly valuable for applications like robotics and sports motion trajectory analytics, where motion trajectory precision is prioritized over visual fidelity.

\end{abstract}    
\section{Introduction}
With the rapid advancement of AI and robotics, predicting robot motion trajectories has become crucial across applications ranging from industrial automation to home services. This capability is vital for ensuring safe, reliable, and efficient robot behavior~\cite{levine2016end, finn2017deep}.
A key challenge in robotic vision prediction is accurately forecasting future scenes based on current visual inputs and action instructions. This enables robots to assess risks, plan ahead, and better understand the interaction between actions and environmental changes—crucial for decision-making and learning.

Our task is to predict what a robot will see 100 frames (10 seconds) into the future, given a current observation image and a text instruction (e.g., “hit the block with the hammer”). This task is challenging as it requires understanding the scene, interpreting the instruction, reasoning about future changes, and generating accurate future frames.

To address this challenge, we implement a deep learning-based multimodal approach that combines the advantages of computer vision, natural language processing, and generative models. Specifically, we fine-tune a pre-trained InstructPix2Pix model~\cite{brooks2023instructpix2pix} (stable diffusion based) to accept a current observation image and a text instruction as input and output a predicted future frame. We use the RoboTwin~\cite{mu2025robotwin} simulation environment to generate training and testing data (data collected in a real-world robotics environment), which provides a simulation platform for various robot interaction tasks.

The main contributions of this article include:

\begin{itemize}
    \item Implementation of a robotic action prediction framework for high-quality future frame generation from current observations and text instructions, with task-specific fine-tuning design of robotic vision generation.

    \item This work establishes the first paradigm for re-architecturing InstructPix2Pix (diffusion-based image editors) into future frame predictors and achieves competent performance. Our design redefines the capabilities of InstructPix2Pix by proposing a multimodal framework that integrates image-text conditioning for future frame prediction. Unlike its original design only for static image editing and modification, unlocking its potential for image forecasting tasks.
 
    \item We conduct our experiments on the real-world RoboTWin dataset, offering greater authenticity and reliability. In the task of predicting future robot actions, our method achieves higher SSIM and PSNR scores compared to existing state-of-the-art video frame prediction methods.

    \item This lightweight design decouples video frame prediction from high computational demands. Common video frame prediction models (e.g., Video Diffusion, Visual Transformers) require high GPU costs and suffer from slow inference, as they need entire video clips as input for high-fidelity frame generation. But in scenarios when motion trajectory accuracy outweighs the need for generating high-fidelity images, our approach provides a more efficient solution—enabling low-cost fine-tuning and inference with just a single image and text instruction, transforming expensive and time-cost video frame prediction into a light image-text multimodal task.
 
\end{itemize}

The remainder of this paper is organized as follows: Section 2 reviews related work; Section 3 introduces the data acquisition method; Section 4 details our methodology; Section 5 presents experimental results and analysis; and finally, Section 6 summarizes this research.
\section{Related Work}

\textbf{Robotic Simulation and Data Generation.}
Training robotic manipulation skills for complex tasks (e.g., dual-arm coordination) requires high-quality demonstration data. While real-world teleoperation provides authentic but scarce samples, Mu et al.~\cite{mu2025robotwin} propose RoboTwin - a hybrid system combining real robot demonstrations with AI-augmented synthetic data. Their key innovation uses real-world task recordings to create digital twins, then employs LLMs to programmatically expand these into diverse training scenarios. This approach maintains physical realism while solving data scarcity.

\textbf{Instruction-Based Image Editing.}
Brooks et al.~\cite{brooks2023instructpix2pix} address the distinct task of editing images directly from human-written instructions. InstructPix2Pix addresses this task by training a diffusion model on a large-scale synthetic dataset composed of GPT-3 instructions paired with edited images from Stable Diffusion. This allows the model to perform diverse edits from natural language commands such as object replacement or style changes. Another related application is Pix2Pix-Zero~\cite{parmar2023pix2pixzero}, similar to InstructPix2Pix, it also focuses on static image editing rather than predictive generation.

\textbf{Robot Position Prediction.}
Accurate robot localization is critical for dynamic logistics. Che et al.~\cite{che2023deep} propose a deep learning solution using a 2D-CNN to predict robot positions from synchronized accelerometer, gyroscope, and magnetometer data. Their method emphasizes rigorous preprocessing and a custom Asymmetric Gaussian loss function to address sensor noise, showcasing improved spatial accuracy. However, this work focuses on predicting the robot's future coordinates, rather than forecasting future action frames (image-based prediction).

\textbf{Video Frame Prediction with Diffusion and Transformer Models.}
Recent works extend diffusion models and visual transformers for video frame prediction. Video Diffusion Models (VDM~\cite{ho2022VDM}, LVD~\cite{yu2023LVD}) and transformer-based methods like VVT~\cite{yan2022videotransformer} or VideoMAE~\cite{tong2022videomae} model temporal consistency to generate realistic video sequences. However, these models require multiple consecutive frames or full video clips as input and are computationally expensive during inference. Additionally, they lack support for multimodal conditioning, such as combining vision and language (Our work only requires 1 frame and 1 instruction as input to do predicting).

\textbf{Multimodal Prediction with Large Models.}
Large-scale multimodal models like Flamingo~\cite{alayrac2022flamingo} and MERLOT~\cite{zellers2022merlot} Reserve integrate image, video, and language understanding via massive pre-training. While effective in few-shot tasks, their huge parameter sizes and GPU memory demands make them impractical for efficient fine-tuning or deployment in real-time robotic systems and other creative real time scenarios~\cite{wu2022nuwa,chen2022mugen}. Their high resource cost limits their applicability in lightweight, fast-inference scenarios like motion prediction from single images.

Unlike the above approaches, our method leverages InstructPix2Pix and fine-tune it in a novel way for robotic frame prediction, combining the benefits of multimodal control and lightweight inference in a unified framework.

\section{Real-World Robotics Data Acquisition}

\subsection{RoboTwin Data Generation}
To facilitate the fine-tuning and evaluation of our model for robotic action frame prediction, we constructed a specialized dataset utilizing the RoboTwin simulation environment~\cite{mu2025robotwin}, which enables realistic robotic interaction data generation based on predefined tasks and instructions. Our data generation pipeline adheres to the project specifications and consists of three primary stages.

First, we focused on three tasks: beat the block with the hammer, handover the blocks and stack blocks. We generated 100 episodes per task, each with 300–500 frames capturing the robot’s perspective and actions during task execution. 

Second, Since only specific visual input was relevant for frame prediction, we extracted RGB images from the robot’s head-mounted camera, excluding depth/non-visual modalities. Extracted frames (minimum $128 \times 128$) were saved in JPG format, reducing data volume while preserving essential visual content.

Third, to align the generated data with our fine-tuning framework, specifically InstructPix2Pix, we organized the images and task prompts into structured sample directories. Each sample includes an initial frame, target frame (100 steps later), and a text instruction (e.g., "handover the blocks"). Files are named consistently (e.g., \texttt{000000\_0.jpg}, \texttt{000000\_1.jpg}, \texttt{prompt.json}).

This structured dataset organization directly facilitates the fine-tuning process of the InstructPix2Pix model for predicting future frames based on the current frame and the provided action command (forming a framework for frame prediction controlled by multimodal text and visual inputs).
In the first stage of experiment, the dataset comprises 300 samples, each consisting of an image pair and a text prompt. In the second stage of experiment, we expand it to 10491 samples, covering a wider range of robot motion trajectories.

\noindent{\bf Ethical Considerations.} 
All data used in this study were generated from the RoboTwin simulation environment, which does not involve real-world robotic production environment, operations or human subjects. Therefore, no ethical review is required for this research.



\subsection{Pre-Evaluation}

Before undertaking task-specific fine-tuning, we evaluated the pre-trained InstructPix2Pix model (\texttt{timbrooks/instruct-pix2pix}) to establish a baseline on our task.
This quantifies the pretrained model's zero-shot capability for predicting future visual states based on the current view and text instruction.

\begin{figure}[H]
    \centering
    \includegraphics[width=0.4\textwidth, height=4.5cm]{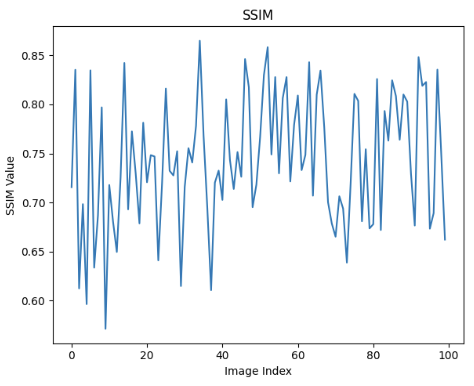}
    \caption{Pre-Evaluation Result - SSIM}
    \label{fig:SSIM}
\end{figure}

\begin{figure}[H]
    \centering
    \includegraphics[width=0.4\textwidth, height=4.5cm]{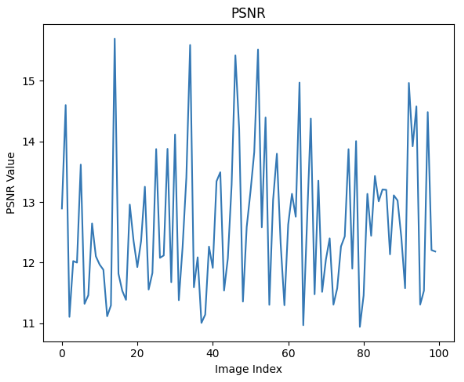}
    \caption{Pre-Evaluation Result - PSNR}
    \label{fig:PSNR}
\end{figure}

The evaluation process utilized a subset of the previously generated dataset, we employed the data corresponding to the "beat the block with the hammer" task. For each of the 100 episodes, we iterated through the sequence of extracted image frames. Pairs of frames were selected as input and ground truth, where the input frame was frame $i$ (\texttt{i.jpg}) and the corresponding ground truth was the frame captured 100 simulation steps later, frame $i+100$ (\texttt{(i+100).jpg}), aligning with the project's prediction objective.

For each selected input frame, we provided it along with the fixed textual instruction "beat the block with the hammer" to the pre-trained InstructPix2Pix pipeline. The model then generated a predicted image for frame $i+100$. Key inference parameters were set as follows: \texttt{num\_inference\_steps=100} and \texttt{image\_guidance\_scale=1}. The generated image was subsequently compared against the actual ground truth image (frame $i+100$) using standard image similarity metrics: Structural Similarity Index (SSIM) and Peak Signal-to-Noise Ratio (PSNR). These metrics were calculated for numerous frame pairs across the episodes, and the results were aggregated (e.g., averaged) to provide a quantitative measure of the original model's performance before any fine-tuning adaptation to the robotic manipulation domain. This baseline is crucial for subsequently assessing the effectiveness of our fine-tuning procedure. From Figure \ref{fig:SSIM} and Figure \ref{fig:PSNR} we can see the pre-trained model performed a SSIM range mostly between 0.65 and 0.85, with PSNR metric ranging largely between 11 and 16.

\section{Methodology}

This section details our implementation of robotic action visual prediction. Aiming at predicting the image the robot will see 100 frames after executing that instruction based on a current observed frame, we fine-tune the pre-trained InstructPix2Pix model \cite{brooks2023instructpix2pix} and design a training strategy adapted to the robotic behavior prediction tasks.

We design a robotic action visual prediction model that takes a current observation image and a textual instruction as input and predicts the corresponding future frame after executing the instruction.

\subsection{Problem Definition}

Given a current observation image $I_t \in \mathbb{R}^{H \times W \times 3}$ and the corresponding text instruction $T$, our goal is to predict the future frame $I_{t+\Delta t} \in \mathbb{R}^{H \times W \times 3}$, where $\Delta t = 100$ frames. Formally, our model $f_\theta$ aims to minimize the difference between the predicted image $\hat{I}_{t+\Delta t} = f_\theta(I_t, T)$ and the actual future frame $I_{t+\Delta t}$:

\begin{equation}
\theta^* = \arg\min_\theta \mathbb{E}_{(I_t, T, I_{t+\Delta t})} [\mathcal{L}(f_\theta(I_t, T), I_{t+\Delta t})]
\end{equation}

where $\mathcal{L}$ is the loss function used to measure prediction quality, and $\theta$ represents the model parameters.

\subsection{Model Architecture Based on InstructPix2Pix}

InstructPix2Pix \cite{brooks2023instructpix2pix} is a latent diffusion-based image editing model that modifies images via text instructions. Although InstructPix2Pix was initially designed for image editing, its architecture is naturally suited for our task, as robotic action prediction is essentially a "temporal edit" of the current observation image.

\begin{figure}[htbp]
\centering
\begin{subfigure}[b]{0.5\textwidth}
    \centering
    \includegraphics[width=\textwidth]{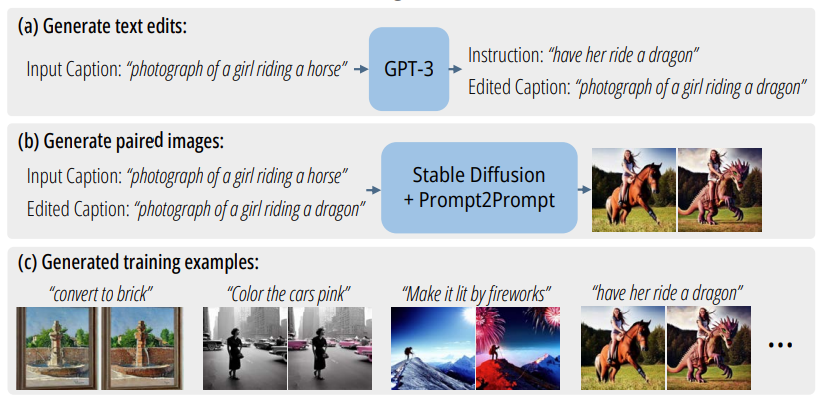}
    \caption{Training data generation process of InstructPix2Pix: (1) Text edit generation with GPT-3, (2) Image pair generation with Stable Diffusion, (3) Building a large-scale training dataset}
    \label{fig:training_data_gen}
\end{subfigure}

\vspace{0.3cm}

\begin{subfigure}[b]{0.5\textwidth}
    \centering
    \includegraphics[width=\textwidth]{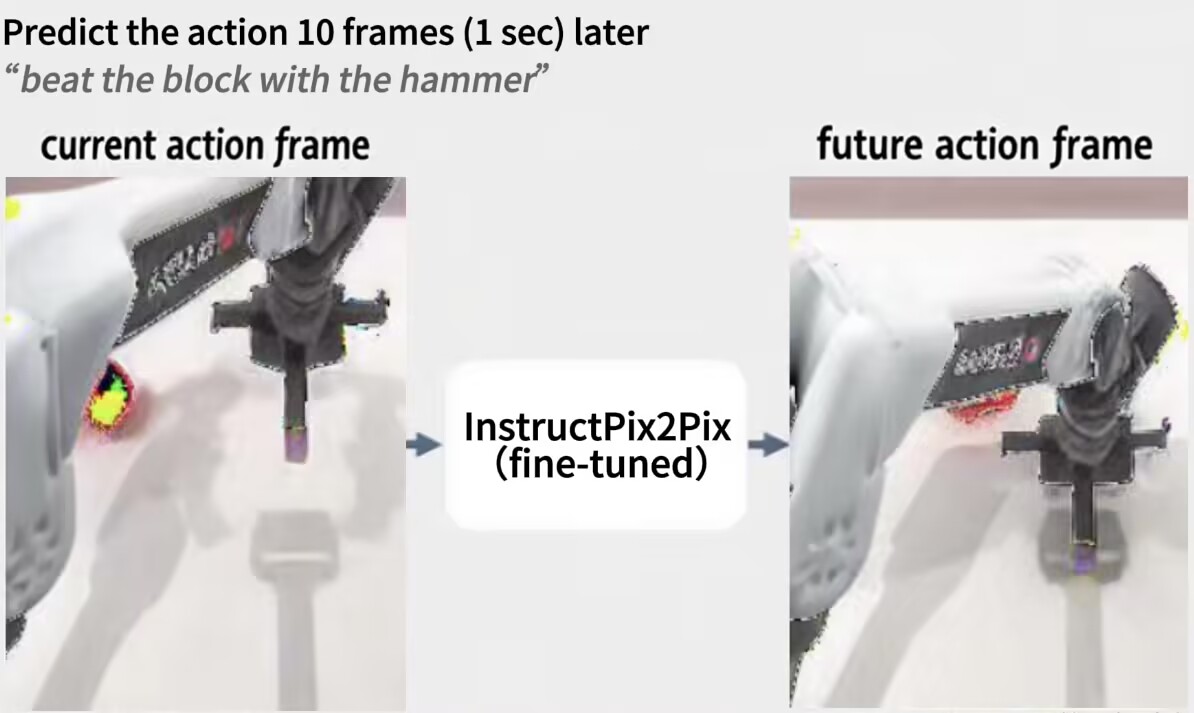}
    \caption{Our designed robotic prediction framework's goal is to predict future action frame based on current observation and the action instruction.\\This paper focuses on the predictive capabilities demonstrated by InstructPix2Pix (fine-tuned), while image fidelity is not the primary focus.}
    \label{fig:instruction_model}
\end{subfigure}
\caption{The top image shows the training data generation process of InstructPix2Pix and the bottom image demonstrates our design - InstructPix2Pix fine-tuned with RoboTwin.}
\label{fig:p2p}
\end{figure}

Figure \ref{fig:p2p} shows the overall workflow of the original InstructPix2Pix method. The approach consists of two main parts: (a) generating text edits using language models, (b) creating paired images based on these text edits, (c) building a large-scale training dataset, and (d) training a diffusion model that can perform edits on real images based on instructions. Figure \ref{fig:p2p} also shows our design for robot action prediction task. In our adaptation, we leverage this framework but modify it to handle temporal prediction rather than just spatial edits. 

Our adapted model architecture mainly includes the following components:

\textbf{Image Encoder:} Converts the input image $I_t$ to a latent representation $z_t$, using VAE (e.g., \texttt{vae-ft-mse})~\cite{rombach2022highresolutionVAE}.  

\textbf{Text Encoder:} Converts the input instruction $T$ into embedding $e_T$ using a pre-trained CLIP~\cite{radford2021clip} encoder.

\textbf{Conditional U-Net:} The core component, which receives noisy latent vectors, image latent representations, and text embeddings, and generates the target latent representation $z_{t+\Delta t}$ through a step-by-step denoising process.

\textbf{Image Decoder:} Decodes the generated latent representation $z_{t+\Delta t}$ into the final predicted image $\hat{I}_{t+\Delta t}$ with a fine-tuned VAE.

\subsection{Model Fine-tuning Strategy}

To adapt the pre-trained InstructPix2Pix model to the robotic action prediction task, we employ the following fine-tuning strategy:

\textbf{Task-specific Conditional Input:} We modify the text input format to combine the robot action instruction with the intent to predict the future, for example: "Scene after executing 'hit the block with the hammer'." This instruction format helps the model understand the task objective.

\textbf{Parameter-Efficient Fine-tuning:} Considering computational resource limitations and to avoid overfitting, we adopt a parameter-efficient fine-tuning (PEFT) approach. Specifically, we freeze most parameters of the pre-trained model and only fine-tune the following components:
\begin{itemize}
    \item Cross-attention layers in the conditional U-Net to enhance the model's understanding of robotic action instructions
    \item Partial parameters of self-attention layers to improve the model's ability to capture spatial relationships
    \item The last few layers of the image encoder to better adapt to visual features in robotic environments
\end{itemize}

\textbf{Progressive Training Strategy:} We employ a two-stage training method: first fine-tuning the model at a lower resolution (64$\times$64), then gradually increasing to the target resolution (128$\times$128). This strategy helps stabilize the training process and improve final performance.

\subsection{Loss Function Design}

Our loss function combines multiple components to ensure the quality of the generated images and consistency with the actual future frames:

\begin{equation}
\mathcal{L} = \lambda_1 \mathcal{L}_{\text{diff}} + \lambda_2 \mathcal{L}_{\text{perc}} + \lambda_3 \mathcal{L}_{\text{adv}}
\end{equation}

We use a composite loss combining (i) latent-space diffusion reconstruction loss (Measures the difference between predicted latent representations and targets), (ii) perceptual loss using VGG features~\cite{johnson2016perceptual} (Calculates the difference between predicted images and actual future frames), and (iii) adversarial loss from a lightweight discriminator to improve realism. We empirically set $\lambda_1=1.0$, $\lambda_2=0.1$, and $\lambda_3=0.01$ to balance their contributions. The formula follows the classifier-free guidance strategy~\cite{ho2021classifier}.

\subsection{Inference Process}
During inference, given a new image-text pair, we encode the inputs and generate the future frame via a conditional denoising process. We adopt classifier-free guidance and 100-step DDIM sampling to balance control and diversity. Lightweight post-processing (color and sharpening) is applied for visual enhancement.

In the inference stage, given a new observation image and text instruction, the model first encodes the image into a latent representation, then combines it with the text embedding, generates the predicted future frame latent representation through a conditional diffusion process, and finally decodes it into the final image. 

To improve the quality and diversity of generated images, we employ the following strategies:

\textbf{Classifier-free Guidance:} Control the degree to which the generation process follows the input conditions by adjusting the weight between conditional and unconditional predictions.
\begin{equation}
\epsilon_\theta(z_t, c) = w \cdot \epsilon_\theta(z_t, c) + (1-w) \cdot \epsilon_\theta(z_t, \emptyset)
\end{equation}

\textbf{Multi-step Sampling:} Use a 100-step DDIM sampling process to balance speed and quality.

\textbf{Post-processing:} Apply lightweight post-processing to enhance the visual quality of generated images, including color balancing and sharpening.

With this approach, our fine-tuned model can generate high-quality, semantically consistent future frame predictions based on current observations and text instructions, providing valuable visual predictions for robotic decision-making.

\section{Experiment}

\subsection{Dataset}
We used the RoboTwin simulator to generate our training and testing dataset. For each sample in the dataset, it contains approximately 400 frames, covering an initial frame, a text instruction, and a ground truth frame 100 steps later. we focus on three specific robotic tasks:
\begin{itemize}
    \item block\_hammer\_beat: The robot beats a block with a hammer (text instruction is "beat the block with the hammer")
    \item block\_handover: The robot performs a handover action with blocks (text instruction is "handover the blocks")
    \item blocks\_stack\_easy: The robot stacks blocks on top of each other (text instruction is "stack blocks")
\end{itemize}

In the first stage of the experiment, for each task, we generated 100 observations, resulting in a total of 300 samples. 
In the second stage of the experiment, we constructed the training set by pairing every 10th frame from each sample (e.g., frame 0 - frame 100, frame 10 - frame 110, ...), ultimately creating a total of 10,491 image pair samples.

\subsection{Implementation Details}
Our model was implemented in PyTorch based on InstructPix2Pix. Fine-tuning used a batch size of 8, AdamW optimizer (lr=1e-4, weight decay=0.01), FP16 training, and a UNet with 320 base channels. We used 1000 diffusion timesteps and [1,2,4,4] multipliers. Attention was applied at resolutions [4,2,1] with 8 heads and 1 transformer block.

We used a pretrained Stable Diffusion v1-5~\cite{rombach2022highresolutionVAE} checkpoint as our starting point and disabled EMA (Exponential Moving Average) during fine-tuning. The scheduler employed a warm-up strategy with LambdaLinear scheduling. Training was performed on a single NVIDIA A100 (40GB). The full 50-epoch training took ~8 hours in stage 1 and ~1 hour per epoch in stage 2 (10k+ samples). Notably, our fine-tuning framework can run on a 24–32GB GPU with reduced batch sizes (batch size = 2 or 1). This is a major advantage of our design. In contrast, conventional multimodal models like Flamingo-3B, used for video frame prediction, require at least 80GB+ of VRAM for training and fine-tuning even with a batch size of 1.

\begin{table}[h]
\centering
\caption{Comparison of Model Parameters and VRAM Requirements for Fine-Tuning}
\label{tab:vram_comparison}
\begin{tabular}{lrr}
\toprule
\textbf{Model} & \textbf{Parameters} & \textbf{VRAM (Fine-Tuning)} \\
\midrule
InstructPix2Pix & $\sim$1.5B & 24--40GB \\
LVD\cite{yu2023LVD} & $\sim$1--3B & 48--80GB+ \\
VDM\cite{ho2022VDM} & $\sim$500M--2B & 32--64GB+ \\
Flamingo-3B\cite{alayrac2022flamingo} & 3B & 80GB+ \\
\bottomrule
\end{tabular}
\end{table}

\subsection{Evaluation Metrics}
To evaluate our model's performance, we used two standard image quality assessment metrics:

\begin{itemize}
    \item Structural Similarity Index (SSIM): Measures the similarity between the predicted future frame and the ground truth based on luminance, contrast, and structure.
    \item Peak Signal-to-Noise Ratio (PSNR): Measures the ratio between the maximum possible power of a signal and the power of corrupting noise that affects the fidelity of its representation.
\end{itemize}

\subsection{Results}
Our model achieves high-quality future frame predictions. In stage 1, SSIM improved from 0.88 to 0.90, and PSNR increased from 36.30 to 38.19 as training epochs increased from 10 to 50. In stage 2 of the experiment, with more diverse samples and broader range of robotic motions, only 2–10 epochs were required to reach comparable and high performance (SSIM: 0.93, PSNR: 39.71).

\begin{table}[h]
\centering
\caption{Quantitative evaluation of our model's performance on test set in the first stage of the experiment.}
\begin{tabular}{|c|c|c|}
\hline
\textbf{Model} & \textbf{SSIM} & \textbf{PSNR (dB)} \\
\hline
10 epochs  & 0.8871 & 36.30 \\
50 epochs  & 0.9094 & 38.19 \\
\hline
\end{tabular}
\label{tab:results1}
\end{table}

\begin{table}[h]
\centering
\caption{Quantitative evaluation in the second stage of the experiment.}
\begin{tabular}{|c|c|c|}
\hline
\textbf{Model} & \textbf{SSIM} & \textbf{PSNR (dB)} \\
\hline
2 epochs  & 0.9066 & 38.04 \\
10 epochs  & 0.9323 & 39.71 \\
\hline
\end{tabular}
\label{tab:results2}
\end{table}
 
Through comparison with other state-of-the-art future frame prediction models, it can be observed that InstructPix2Pix, after being fine-tuned by the RoboTwin framework, demonstrates exceptional performance in the field of robotic motion trajectory prediction.

\begin{table}[H]
\centering
\caption{Performance Comparison of Frame Prediction Models, InstructPix2Pix (FT) is our fine-tuned model.}
\label{tab:model_performance}
\begin{tabular}{l l c c}
\toprule
\textbf{Model} & \textbf{Dataset} & \textbf{SSIM} & \textbf{PSNR} \\
\midrule
ConvLSTM\cite{shi2015convlstm} & Moving MNIST & 0.75 & 28.5 \\
VDM & UCF-101 & 0.87 & 35.7 \\
LVD & Kinetics-600 & 0.89 & 36.5 \\
Flamingo-3B & SSv2 & 0.72 & 28.3 \\
MAGVITv2\cite{yu2023magvitv2} & BAIR & 0.91 & 37.2 \\
MCVD\cite{voleti2022mcvd} & RoboNet & 0.89 & 36.8 \\
InstructPix2Pix (FT) & RoboTwin & 0.93 & 39.7 \\
\bottomrule
\end{tabular}
\end{table}

\subsection{Qualitative Analysis}
\begin{figure}[H]
\centering
\includegraphics[width=0.3\textwidth, height=4.55cm]{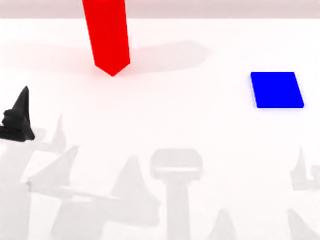}
\caption{Input Image}
\label{fig:input_image}
\end{figure}

\begin{figure}[H]
\centering
\includegraphics[width=0.3\textwidth, height=4.55cm]{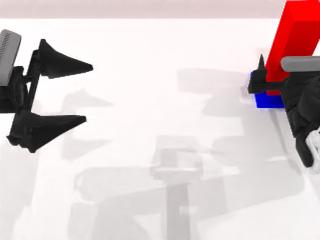}
\caption{Ground Truth Image}
\label{fig:groundTruth}
\end{figure}

\begin{figure}[H]
\centering
\includegraphics[width=0.3\textwidth, height=4.55cm]{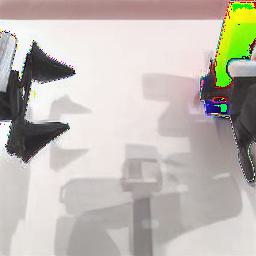}
\caption{Predicted Output}
\label{fig:predicted_output}
\end{figure}

Figure~\ref{fig:input_image}–\ref{fig:predicted_output} show examples of our model's predictions compared to the ground truth images for the robotic tasks. Our model successfully captures motion transformations across tasks, producing visually consistent results.

\vspace{1cm}

With more training samples and a wider range of action coverage in the second stage of the experiment, we also implement multi-frame prediction, predicting consecutive multiple frames in the future while maintaining superior performance metrics.

\begin{figure*}[htbp]
    \centering
    \begin{subfigure}{0.3\textwidth}
        \includegraphics[width=\textwidth, height=4cm]{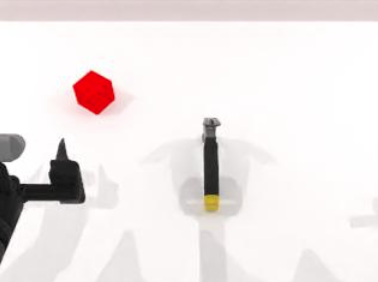}
        \caption{Input Frame (Frame 47)}
        \label{fig:input}
    \end{subfigure}
    
    \centering
    \begin{subfigure}{0.3\textwidth}
        \includegraphics[width=\textwidth, height=4cm]{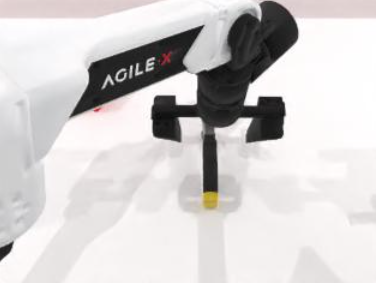}
        \caption{Ground Truth Frame 147}
        \label{fig:gt1}
    \end{subfigure}
    \hfill
    \begin{subfigure}{0.3\textwidth}
        \includegraphics[width=\textwidth, height=4cm]{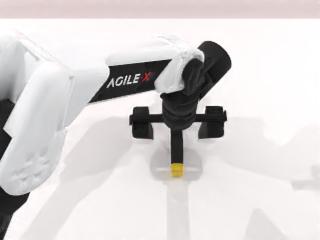}
        \caption{Ground Truth Frame 148}
        \label{fig:gt2}
    \end{subfigure}
    \hfill
    \begin{subfigure}{0.3\textwidth}
        \includegraphics[width=\textwidth, height=4cm]{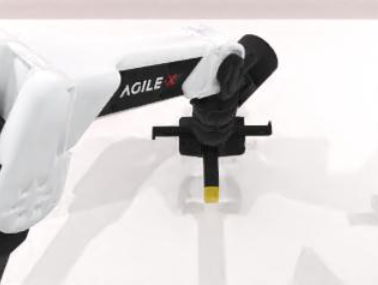}
        \caption{Ground Truth Frame 149}
        \label{fig:gt3}
    \end{subfigure}
    
    \centering
    \begin{subfigure}{0.3\textwidth}
        \includegraphics[width=\textwidth, height=4cm]{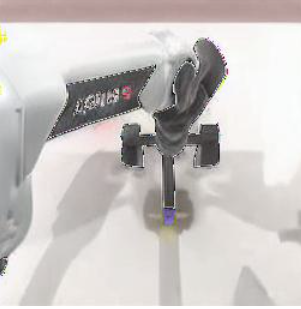}
        \caption{Predicted Frame 147}
        \label{fig:pred1}
    \end{subfigure}
    \hfill
    \begin{subfigure}{0.3\textwidth}
        \includegraphics[width=\textwidth, height=4cm]{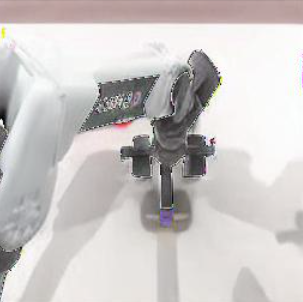}
        \caption{Predicted Frame 148}
        \label{fig:pred2}
    \end{subfigure}
    \hfill
    \begin{subfigure}{0.3\textwidth}
        \includegraphics[width=\textwidth, height=4cm]{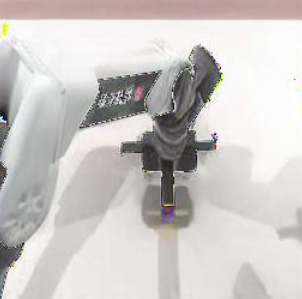}
        \caption{Predicted Frame 149}
        \label{fig:pred3}
    \end{subfigure}
    
    \caption{Input, Ground Truth and multiple Predicted Frames (Within 10 epochs training)\\In scenarios
such as robotic movement trajectory, precise predictive capability of
motion trajectories is our primary focus (emergence capability in our research), while image resolution is not the foremost priority in this study.}
    \label{fig:frame_comparison}
\end{figure*}

Figure~\ref{fig:frame_comparison} illustrates the predicted consecutive multiple frames and consecutive ground truth images (we randomly selected the 47th frame as the input, and we could see the predicted results after 100 frames (10 seconds)).
The results demonstrate that our approach generates highly accurate action trajectory predictions, with the forecasted motion closely aligning with the ground truth (In scenarios such as robotic movement trajectory, precise prediction of motion trajectories is our primary concern, while image fidelity is not the foremost priority). Experimental validation confirms that robust trajectory prediction with our framework can be achieved with just 10,000 training pairs and only 2-10 epochs of training.

\vspace{1cm}

\subsection{Inference Time for Predicting Future Frames}  

In our experiment on an NVIDIA A100 (40GB), our fine-tuned InstructPix2Pix generates 15 future frames in about \textbf{38 seconds}. For reference, large multimodal baselines typically require minutes under similar conditions: Flamingo-3B takes over 12 minutes, while LVD needs around 8--10 minutes. This highlights the efficiency advantage of our lightweight adaptation for robotic prediction tasks.
Even without specific inference optimizations, our framework shows clear efficiency advantages for robotic prediction tasks, with further opportunities for speedup via quantization and other acceleration techniques.

\subsection{Task-specific Analysis}
We further analyzed our model's performance on each individual task to understand its strengths and limitations:

\begin{itemize}
    \item \textbf{Block hammer beat}: The model accurately predicted the position of the hammer and its interaction with the block. The predicted frames correctly captured the spatial relationships between objects and the motion blur associated with the hammering action.
    
    \item \textbf{Block handover}: This task involves more complex hand-object interactions. Our model successfully predicted the general movement and positioning of the blocks during handover, though with slightly less precision in finger positions compared to the ground truth.
    
    \item \textbf{Blocks stacking}: The model performed exceptionally well on this task, accurately predicting the final stacked configuration of blocks with proper shadowing and perspective.
\end{itemize}

Overall, the model demonstrated robust performance across all three robotic tasks, and it illustrates great potential in other action prediction tasks.

\subsection{Training Progress}
\begin{figure}[H]
\centering
\begin{subfigure}[b]{0.48\textwidth}
\includegraphics[width=\textwidth]{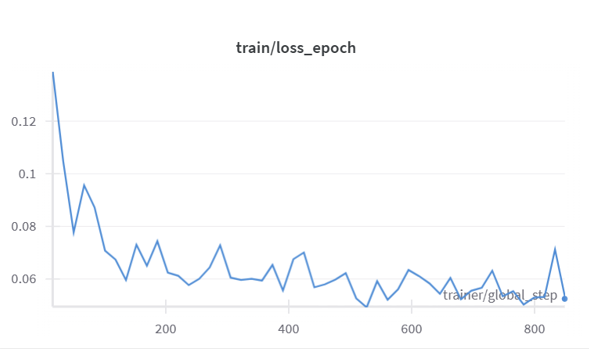}
\caption{Training Loss}
\end{subfigure}
\hfill
\begin{subfigure}[b]{0.48\textwidth}
\includegraphics[width=\textwidth]{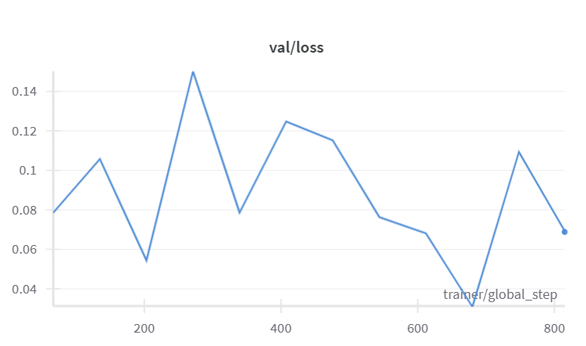}
\caption{Validation Loss}
\end{subfigure}
\caption{Training and validation loss curves.}
\label{fig:training_curves}
\end{figure}

Figure \ref{fig:training_curves} shows the training and validation loss curves over the 50 epochs of training. The consistent decrease in both training and validation loss demonstrates that our model effectively learned to predict future frames without overfitting.

\subsection{Ablation Study}
To understand the contribution of different components in our approach, we conducted a simple ablation study by varying the number of training epochs while keeping other hyperparameters constant. When increasing the number of epochs from 2 to 10 led to significant improvements in both SSIM and PSNR. This highlights the importance of sufficient training iterations for the model to capture the nuances of robotic movements and accurately predict future frames. In our second stage of the experiment, we trained for 10 epochs, and we believe that as the number of epochs continues to increase, we anticipate seeing continued performance gains.
\section{Conclusion}

In this paper, we present a deep learning framework for robotic action visual prediction. By fine-tuning the InstructPix2Pix model on RoboTwin simulation data and create a new framework, our method predicts the robot’s future observation frames from a current action frame and a textual instruction.

This design transforms the computationally expensive video frame prediction task into a more lightweight and controllable multimodal vision-language controlled prediction task. This approach not only improves operational efficiency but also achieves fine-tuning and future frame inference with significantly reduced GPU resource requirements. Furthermore, our experimental results demonstrate the effectiveness of this approach, achieving high-quality predictions with SSIM values of 0.93 and PSNR of 39.71dB in robot action prediction task.

This implementation successfully captures the essential spatial transformations and object relationships in various robotic manipulation scenarios. The model performed particularly well on the robot action tasks, while also showing strong potential in other tasks like football trajectory or other sport trajectory prediction. These results suggest that fine-tuned generative models like  InstructPix2Pix can effectively learn to predict the visual outcomes of robotic actions based on current observations and textual instructions. Moreover, the significantly lower training and inference cost, along with ultra-low inference latency, makes this design possible for real-world deployment and real-time critical applications.

\newpage
{
    \small
    \bibliographystyle{ieeetr}  
    \bibliography{main}
}


\end{document}